\documentclass[11pt]{article}
\usepackage{geometry}
\usepackage{coling2020}
\usepackage{times}
\usepackage{url}
\usepackage{latexsym}
\usepackage{microtype}
\usepackage{amsmath}

\usepackage{url}
\usepackage{breqn}
\usepackage{graphicx}
\usepackage{xspace}
\usepackage{subfigure}
\usepackage{xcolor}
\usepackage{caption}
\usepackage{booktabs}

\newcommand{\ie}{\emph{i.e.,}\xspace}

\usepackage[normalem]{ulem}

\newcommand{\method}{BabelEnconding\xspace}

\colingfinalcopy 

\title{{\method} at SemEval-2020 Task 3: Contextual Similarity as a Combination of Multilingualism and Language Models}

\author{Lucas R. C. Pessutto \\
  Inst. of Informatics \\ UFRGS -- Brazil \\
 {\tt \footnotesize lrcpessutto@inf.ufrgs.br} \\\And
  Tiago de Melo \\
 Amazonas State Univ. \\ UEA -- Brazil\\
{\tt \footnotesize tmelo@uea.edu.br}\\\And
 Viviane P. Moreira \\
 Inst. of Informatics \\ UFRGS -- Brazil\\
{\tt \footnotesize viviane@inf.ufrgs.br}\\ \And
  Altigran da Silva \\
   Federal Univ. of Amazonas \\ UFAM -- Brazil\\
 {\tt \footnotesize alti@icomp.ufam.edu.br}
}
  
\date{}

\begin{document}
\maketitle
\begin{abstract}
  This paper describes the system submitted by our team (\method) to SemEval-2020 Task~3: Predicting the Graded Effect of Context in Word Similarity. 
  We propose an approach that relies on translation and multilingual language models in order to compute the contextual similarity between pairs of words. 
  Our hypothesis is that evidence from additional languages can leverage the correlation with the human generated scores.
  \method was applied to both subtasks and ranked among the top-3 in six out of eight task/language combinations and was the highest scoring system three times.

\end{abstract}

\section{Introduction}
%
%
\blfootnote{
%
%
%
 \hspace{-0.65cm}  
     This work is licensed under a Creative Commons 
     Attribution 4.0 International License.
     License details:
     \url{http://creativecommons.org/licenses/by/4.0/}.
}

Word similarity is a key task in Natural Language Processing (NLP) applications. 
Language models, such as word embeddings~\cite{mikolov2013} create vector representations for the words that are able to capture syntactic and semantic  relationships.
These representations became very popular in the last few years as they have boosted the performance of several NLP tasks.
However, since each word is represented by a fixed vector these techniques have problems dealing with polysemous words and identifying subtle meaning changes between different sentences.
On the other hand, state-of-the-art language models, like BERT \cite{devlin2019} provide a contextualized word representation -- the representation of a word relies on its context, which means that the same word may have different representations through the sentences.
Thus, BERT models are more suitable for handling polysemous words.

Task 3 in SemEval 2020 -- \textit{Predicting the Graded Effect of Context in Word Similarity} \cite{task3description} was motivated by this improvement on language models. 
The task aims at the design of a similarity measure which captures the human perception of the meaning of words.
For that purpose, task organizers built and annotated datasets in four languages -- English, Croatian, Finnish, and Slovenian. 
Each entry in a dataset consists of two target words and two contexts, where each one is a piece of text containing both target words. 
The global task is divided into two subtasks: 1) predicting the change in the human annotator's scores of similarity when presented with the same pair of words within two different contexts; and 2) predicting the human scores of similarity for a pair of words within two different contexts. 

In this paper, we describe {\method}, an approach that relies on machine translation and multilingual language models to evaluate the contextual similarity of pairs of words.
Our hypothesis is that having similarity information from more languages helps decide on how similar the words are.

Considering the eight combinations of language/subtask, \method was ranked among the top-3 competitors six times, and was the top scoring method in three cases.
Our additional experiments in English and Croatian showed that adding more languages noticeably improved the results for Croatian in both subtasks.
In English, the gain was small and happened only in Subtask~2.

\section{Background and Related Work}


The Distributional Hypothesis \cite{harris1954} states that the meaning of a word changes depending on the context it is used. At the same time, this hypothesis also states that if two words tend to be used in the same contexts, then they are likely to be more similar.
This claim inspired many solutions in NLP that based solely on the distribution of words in the corpus~\cite{fernandez2016distributional,wang2020survey,luddecke2019distributional}.
Word embeddings and language models, for example, are among these solutions. 
The idea is to represent words in a vector space in such a way that the semantic similarity between words is preserved.
In the past few years, techniques to build language models became very popular. 
Word2vec~\cite{mikolov2013} is an efficient and fast training method for word embeddings, based on co-occurrence statistics. The authors devised two model architectures for the word vectors training -- continuous bag of words and skip-grams. Both approaches consist of neural networks trained to predict neighbor contextual words. 
Despite its ability in mapping linguistic regularities present in documents, this language model produces a unique representation for each word in the vocabulary, which prevents the differentiation of word senses.

In state-of-the-art language models, such as 
BERT~\cite{devlin2019}, the context of a word is taken into account in its representation. 
These models are trained over a large corpus to predict missing tokens which are removed from the original sentences.
An advantage of BERT over Word2vec is that it creates different representations for the same word depending on the context in which the word appears. Another advantage of BERT-like models is that they can be specialized for a specific task with few training epochs.

Solutions for measuring contextual similarity between word pairs and word-sense disambiguation benefited from BERT-like language models. 
Enriched models were designed~\cite{levine2019sensebert,peters2019knowledge,scarlini2020}, and new datasets  such as the Word-in-Context Dataset~\cite{pilehvar2018wic} and CoSimLex \cite{ArmendarizEtAl20LREC} were assembled.
Word-sense disambiguation can also take advantage of multilingualism.
Some works have employed parallel/comparable corpora~\cite{banea2011,dandala2013} and translation~\cite{carpuat2013} to that task.
Multilingual resources, such as Multi-SimLex~\cite{vulic2020multi}, were also developed and yielded improvements compared with
the monolingual version.



\section{\method}





\begin{figure*}[b]
    \centering
    \includegraphics[width=0.95\textwidth]{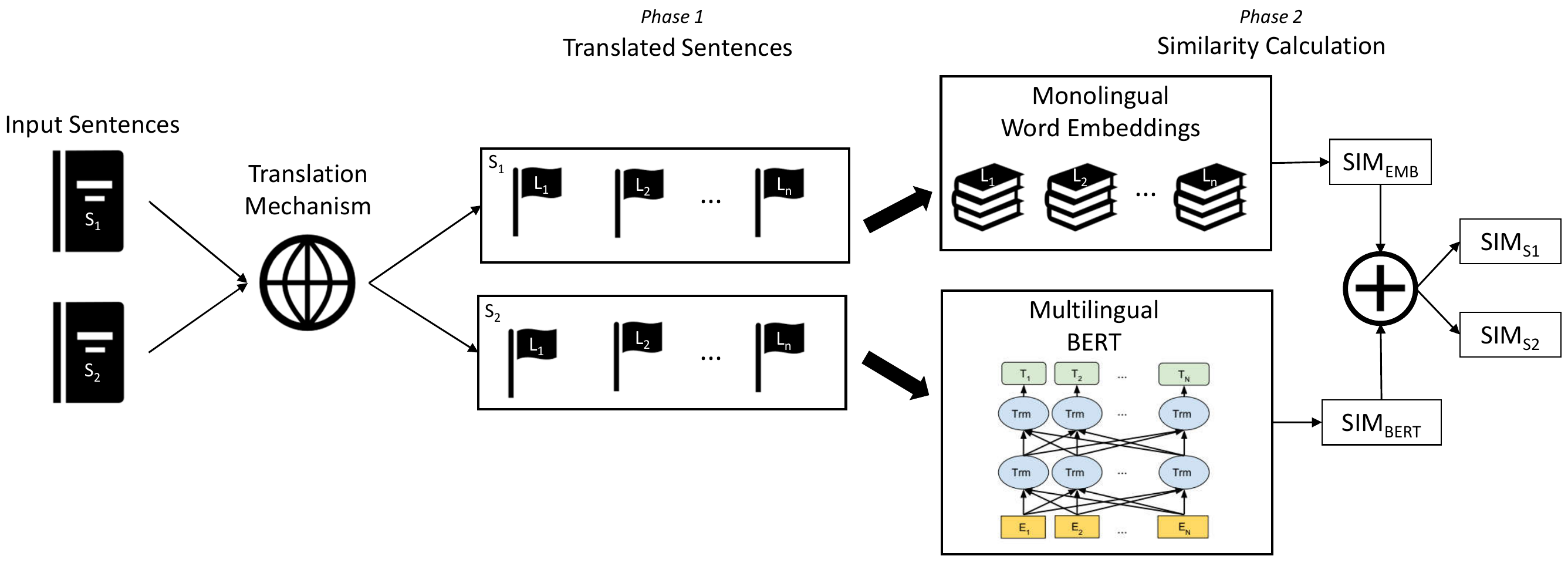}
    \caption{Overview of \method}
    \label{fig:overview}
\end{figure*}

Our proposed solution, called \method, works in two phases and its overall process is depicted in Figure~\ref{fig:overview}. 
The input is a pair of words and two sentences (contexts) containing both words of interest. 
More formally, let $S_1 = \{w^1_1, w^1_2, \ldots, w^1_i\}$ and $S_2 = \{w^2_1, w^2_2, \ldots, w^2_j\}$ be two sentences, where there is a pair of words $p = \langle w_a, w_b \rangle \in S_1$ and $S_2$. 
For example, let $S_1 =$ \textit{``Her prison \textbf{cell} was almost an improvement over her \textbf{room} at the last hostel"} and $S_2 = $ \textit{``His job didn't leave much \textbf{room} for a personal life. He knew much more about human \textbf{cells} than about human feelings"} be two sentences, where the pair of words $p = \langle room, cell\rangle$.

In the first phase of \method, both input sentences $S_1$ and $S_2$ are translated into a set of $k$ languages $L = \{l_1, l_2, \ldots, l_k\}$.
This process will produce a set of translated sentences $S^{l_i} = \left \{S_1^{l_i}, S_2^{l_i} \right \}$, which corresponds to the translation of the original sentences, into each language $l_i$ $\in L$. Then, the words of interest are identified in the translated text, generating two sets $p_{s_1}^{l_i}$ and $p_{s_2}^{l_i}$.
In this example, considering $L = \{\text{Italian}, \text{Portuguese}\}$, the pairs of words of interest are translated as $p_{s_1}^{\text{IT}} = \langle \text{cella}, \text{stanza} \rangle$, 
$p_{s_1}^{\text{PT}} = \langle \text{cela}, \text{quarto} \rangle$
from $S_1$ and 
$p_{s_2}^{\text{IT}} = \langle \text{spazio}, \text{cellule} \rangle$, 
$p_{s_2}^{\text{PT}} = \langle \text{espaço}, \text{células} \rangle$ from $S_2$.


In the second phase, with the translated sentences, we evaluate the similarity between the pair of words in $p$ for each language in $L$ separately in two ways: ($i$) using word embeddings and ($ii$) BERT. 
Finally, \method calculates a weighted average between word embeddings and BERT similarities. These similarities are used to address both subtasks.

\vspace{0.2cm}
\noindent\textbf{Word Embedding Similarity} consists in taking the cosine similarity between the word vectors of the two words in each language. We rely on pre-trained monolingual word embeddings to represent the words. The context is not used in this similarity measure since there is a fixed vector for each word. 

\vspace{0.2cm}
\noindent\textbf{BERT Similarity} requires inferring the word embedding representation of words in BERT models, taking context into consideration. The context is the sentence ($S_m$) containing the two words.
This process was done summing the last four hidden layers of the BERT model. This choice was made based on the good results achieved by \newcite{devlin2019} in the Named Entity Recognition task.

\vspace{0.2cm}
\noindent\textbf{\method Similarity}
consists on a weighted average between word embedding and BERT similarities scaled in multiple languages.  Equation~\ref{equ:sim} shows how \method calculates the similarity between words $w_1$ and $w_2$ within sentence $S_m$. In this equation, $\alpha$ and $\beta$ are the weights given to BERT and Word Embedding similarities, respectively. 
\begin{dmath}
\label{equ:sim}
SIM_{(w_1, w_2)}^{S_m} = \frac{1}{L} \sum_{i = 1}^{L} \alpha SIM_{BERT}(w_1^i, w_2^i, S_m) + \beta SIM_{WE}(w_1^i, w_2^i)
\end{dmath}

Our hypothesis is that having similarity information from more languages helps decide on how similar they are.
The underlying assumption is that if two words are translated to the same word in other language, they are more likely to be more similar. 
Translation also helps identifying dissimilarity between words as it can help to disambiguate terms.

Preliminary tests showed that, once both words occur together in the same context, the similarity between words tended to be undesirably high when using just BERT representations.  
This effect can be attributed to BERT's attention mechanism. 
Thus, a combination of BERT and fixed word embeddings was designed to alleviate this issue.

\section{Experimental Setup}
\label{sec:exp}

\noindent\textbf{Dataset}. The dataset used in our experiments was CoSimLex \cite{ArmendarizEtAl20LREC} which consists of 340 sentence pairs in English (EN), 112 in Croatian (HR), 111 in Slovene (SL), and 24 in Finnish (FI). 
Please refer to that paper for details on the annotation methodology.

\vspace{0.2cm}
\noindent\textbf{Languages}. The source sentences were translated into the following languages: English (EN), Spanish (ES), Italian (IT), Bosnian (BS), German (DE), Greek (EL), Polish (PL), Portuguese (PT), Russian (RU), Serbian (SR), and Turkish (TR). This choice was made based on the main languages used in Word Sense Disambiguation tasks~\cite{camacho2016nasari,duong2017multilingual,resnik2004exploiting,raganato2017neural,fernandez2017word}.






\vspace{0.2cm}
\noindent\textbf{Tools and Resources}.
The official experiments used Google Translator API\footnote{\url{https://cloud.google.com/translate/}}.
Here, we also report a comparison with Bing Microsoft Translator\footnote{\url{https://www.bing.com/translator}}.
The multilingual uncased version of BERT\footnote{\url{https://github.com/google-research/bert/blob/master/multilingual.md}} trained on Wikipedias in 102 languages was used.
For word embeddings, we used FastText\footnote{\url{https://fasttext.cc/docs/en/pretrained-vectors.html}} which provides pre-trained embeddings for 157 languages. These embeddings were also trained on Wikipedia.

\vspace{0.2cm}
\noindent\textbf{Evaluation Metrics}.
The evaluation metrics used to assess the quality of the participating systems measure the correlation between the scores assigned by human annotators and the scores automatically generated by the participating systems (higher scores are better).
In Subtask~1, the uncentered Pearson correlation between was used and, in Subtask~2, the harmonic mean between Pearson and Spearman correlations was used.

\vspace{0.2cm}
\noindent\textbf{Experimental Runs}. 
In order to have a broader evaluation of \method, we tested different system configurations and parameters. 
With the additional runs the goal was answering three questions -- ($i$) \textit{How much each component of \method contributes to the overall result?}; 
($ii$) \textit{Do results improve as more languages are added?}; and ($iii$) \textit{Does  the  translation  mechanism  impact  the  results?}.

\section{Results}

\vspace{0.2cm}
\noindent\textbf{Results for the Official Runs. }
The system configurations that achieved the best results in the official runs are shown in Table~\ref{tab:official_results}.
We varied the number of extra languages and the values for $\alpha$ and $\beta$.
For Subtask~1, English and Slovenian performed better when no additional languages were used in the similarity computation. 
On the other hand, Croatian and Finnish performed better when all 11 additional languages were used. 
Moreover, these two languages were benefited when word embeddings were completely removed from \method calculation.
In Subtask~2, the use of all extra languages or a subset of the 11 languages showed the best results. A combination of BERT and word embeddings also proved to be beneficial for that task.
In comparison with other participants, we achieved best results for Croatian, in both subtasks, and for Slovenian in Subtask~2.  

\begin{table}[tb]
\centering
\caption{Official results for \method. Numbers in brackets indicate our ranking.}
\label{tab:official_results}
\resizebox{\textwidth}{!}{%
\begin{tabular}{lcc|llll|llll}
\toprule
\multicolumn{3}{c|}{Configuration} & \multicolumn{4}{c|}{Correlation scores Subtask~1}    
& \multicolumn{4}{c}{Correlation scores Subtask~2}                             \\
Extra Langs.  & $\alpha$    & $\beta$    & EN             & HR             & FI             & SL             & EN             & HR             & FI             & SL             \\ \hline
None               & 0.7      & 0.3     & \textbf{0.730}$^{(7)}$ & 0.634          & 0.607          & \textbf{0.646}$^{(3)}$ & 0.615          & 0.583          & 0.376          & 0.559          \\

PT, EL, TR, RU     & 0.8      & 0.2     & 0.683          & 0.703          & 0.707          & 0.617          & 0.620          & 0.635          & \textbf{0.611}$^{(2)}$ & 0.578          \\
ES, IT, PT, DE     & 0.6      & 0.4     & 0.709          & 0.735          & \textbf{0.726}$^{(3)}$ & 0.525          & 0.626          & 0.647          & 0.571          & \textbf{0.579}$^{(1)}$ \\
11 Languages       & 0.7      & 0.3     & 0.695          & 0.716          & 0.718          & 0.575          & \textbf{0.634}$^{(10)}$ & \textbf{0.658}$^{(1)}$ & 0.581          & 0.566          \\
11 Languages       & 1.0      & 0.0     & 0.711          & \textbf{0.740}$^{(1)}$ & \textbf{0.726}$^{(3)}$ & 0.624          & 0.614          & 0.632          & 0.557          & 0.578          \\
\hline
\multicolumn{3}{c|}{Best Score in SemEval Task3} & 0.774 & 0.740 & 0.772 & 0.654 & 0.723 & 0.658 & 0.645 & 0.579 \\
\bottomrule
\end{tabular}
}
\end{table}

Table~\ref{tab:team-results} summarizes the official results for both Subtask~1 and Subtask~2. The column \textit{Average} shows the average of the results achieved by the teams among all languages and the column \textit{Rank} shows the team's position in the ranking. As we can see, our method performed well in both subtasks, being ranked in first place considering the average of all languages.

\begin{table}[h]
\centering
\caption{Official Results for Subtask~1 and Subtask~2 for all participating teams. Bold indicates the best result for the given language. \cite{task3description}}
\label{tab:team-results}
\scalebox{0.83}{
\begin{tabular}{l|c|c|c|c|c|c|c|c|c|c|c|c}
\toprule
              & \multicolumn{6}{c|}{Subtask~1}                    & \multicolumn{6}{c}{Subtask~2}                    \\ \midrule
TEAM          & EN    & HR    & FI    & SL    & Average & Rank & EN    & HR    & FI    & SL    & Average & Rank \\ \hline
BabelEncoding & 0.730  & \textbf{0.740} & 0.726 & \textbf{0.646} & \textbf{0.710}  & 1    & 0.634 & \textbf{0.658} & 0.611 & \textbf{0.579} & \textbf{0.620}  & 1    \\ \hline
Team 1        & \textbf{0.774} & 0.634 & 0.745 & 0.605 & 0.689  & 2    & 0.437 & 0.397 & 0.357 & 0.345 & 0.384   & 10   \\ \hline
Team 2        & 0.768 & 0.594 & \textbf{0.772} & 0.583 & 0.679 & 3    & 0.695 & 0.385 & 0.341 & 0.485 & 0.476  & 6    \\ \hline
Team 3        & 0.754 & 0.664 & 0.626 & 0.648 & 0.673   & 4    & 0.715 & 0.545 & \textbf{0.645} & 0.573 & 0.619  & 2    \\ \hline
Team 4        & 0.712 & 0.681 & 0.574 & \textbf{0.654} & 0.655 & 5    & 0.695 & 0.616 & 0.255 & 0.510  & 0.519   & 5    \\ \hline
Team 5        & 0.754 & 0.616 & 0.360  & 0.560  & 0.572  & 6    & 0.720  & 0.565 & 0.354 & 0.483 & 0.530  & 4    \\ \hline
Team 6        & 0.738 & 0.440  & 0.546 & 0.512 & 0.559   & 7    & -     & -     & -     & -     & -       & -    \\ \hline
Team 7        & 0.529 & 0.531 & 0.399 & 0.510  & 0.492 & 8    & -     & -     & -     & -     & -       & -    \\ \hline
Team 8        & 0.042 & 0.587 & 0.671 & 0.603 & 0.475 & 9    & 0.647 & 0.402 & 0.289 & 0.516 & 0.463  & 7    \\ \hline
Team 9        & 0.721 & 0.416 & 0.025 & 0.624 & 0.446  & 10   & -     & -     & -     & -     & -       & -    \\ \hline
Team 10       & 0.544 & 0.374 & 0.389 & 0.328 & 0.408 & 11   & \textbf{0.723} & 0.613 & 0.597 & 0.487 & 0.605   & 3    \\ \hline
Team 11       & -     & -     & -     & -     & -       & -    & 0.573 & 0.402 & 0.289 & 0.516 & 0.445   & 8    \\ \hline
Team 12       & -     & -     & -     & -     & -       & -    & 0.340  & 0.338 & 0.454 & 0.411 & 0.385 & 9    \\ \bottomrule
\end{tabular}
}
\end{table}

\begin{figure}[htb]
\centering
\includegraphics[width=0.30\linewidth]{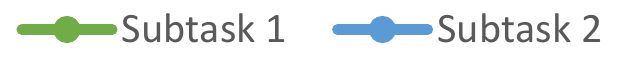}\\
\subfigure[English]{\includegraphics[width=0.49\linewidth]{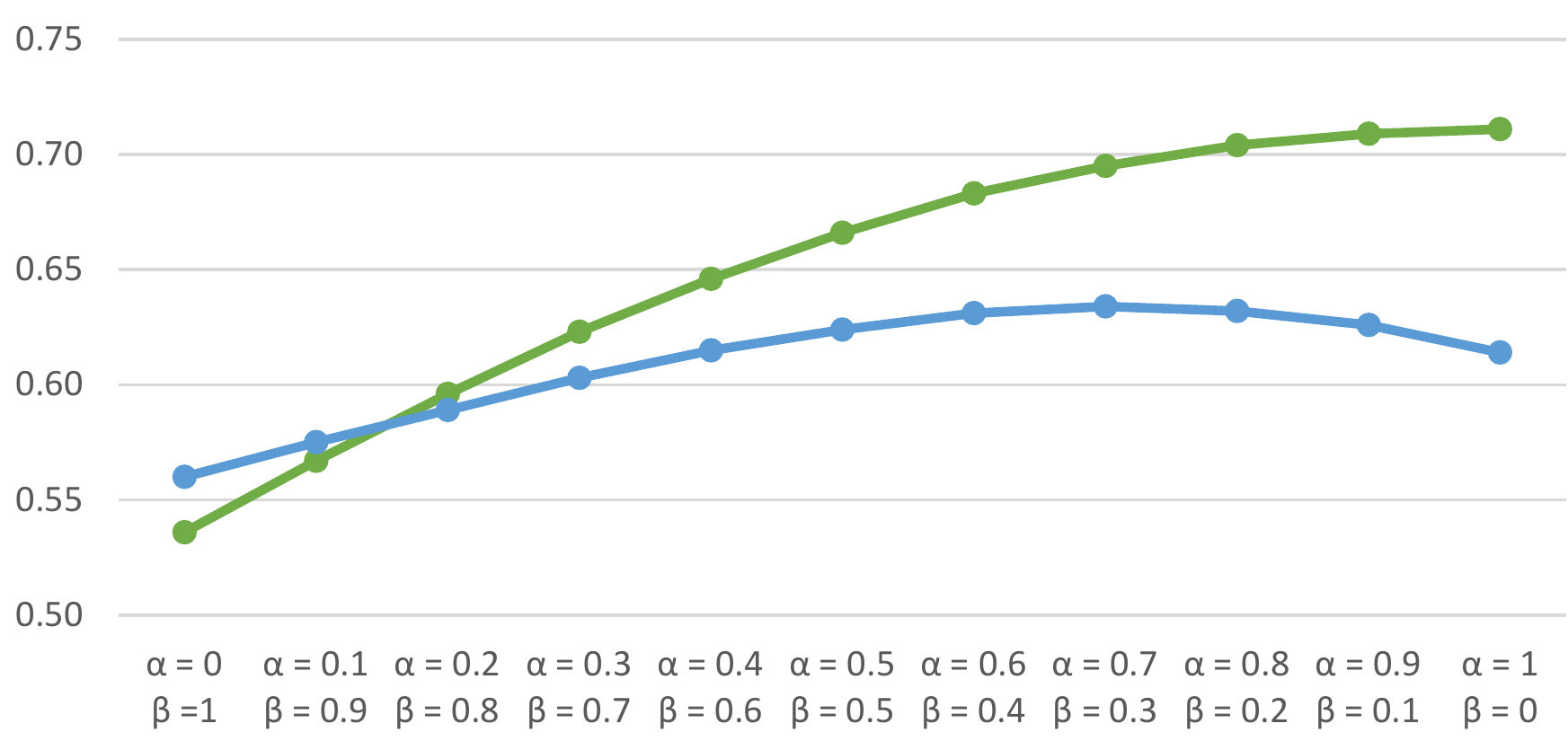}}
\subfigure[Finnish]{\includegraphics[width=0.49\linewidth]{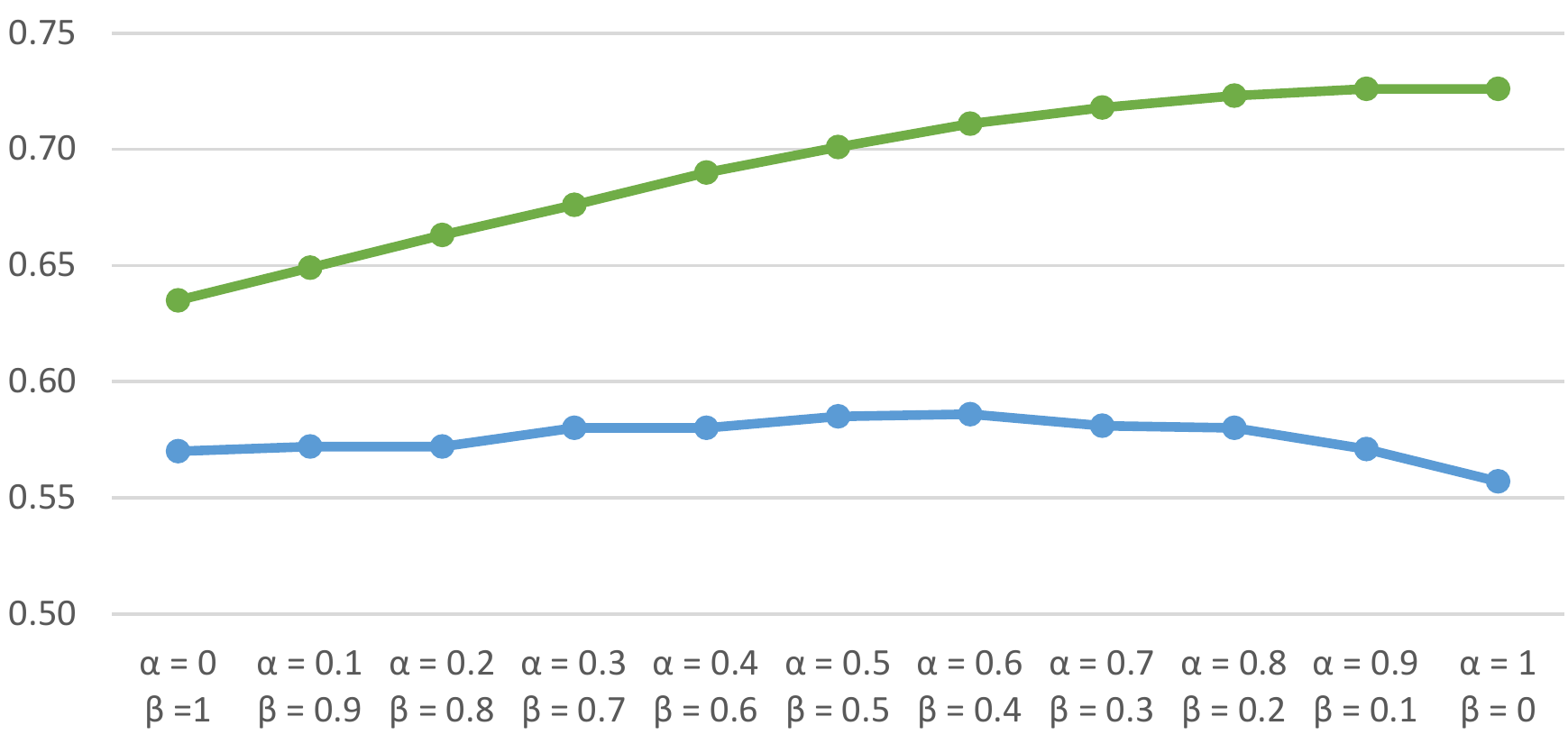}}\\
\caption{Correlation scores with human judges for different values of $\alpha$ and $\beta$ in \method}
\label{fig:alfa_beta}
\end{figure}

\vspace{0.3cm}
\noindent\textbf{How much each component of \method contributes to the overall result? }
In order to assess the contribution of the components of \method, we performed experiments varying the parameters $\alpha$ (which scales the contribution of BERT similarity), and $\beta$ (which weighs the importance of word embedding similarity). 
As a general tendency, increasing $\alpha$ values tends to produce better correlation results, especially in Subtask~1. 
However, when the word embeddings component is removed (\ie $\beta$=0), results tend to get worse, mainly in Subtask~2.
Figure~\ref{fig:alfa_beta} shows the results for English and Finnish. 
The curves in (a) represent the typical case, which was found in English, Croatian, and Slovenian. 
The results for Finnish (b) in Subtask~2 followed a different pattern, in which evaluation scores are not affected by the presence of BERT on similarity computation.
We believe this happened because Finnish is an agglutinative language, and since BERT's tokenization process uses Byte Pair Encoding, it tends to split Finnish words in too many tokens \cite{virtanen2019} yielding to poorer word representations.


\begin{figure}[bht]
\centering
{\includegraphics[width=0.75\linewidth]{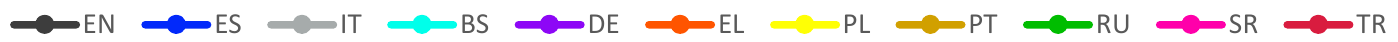}}\\
\subfigure[EN - Subtask~1]{\includegraphics[width=0.49\linewidth]{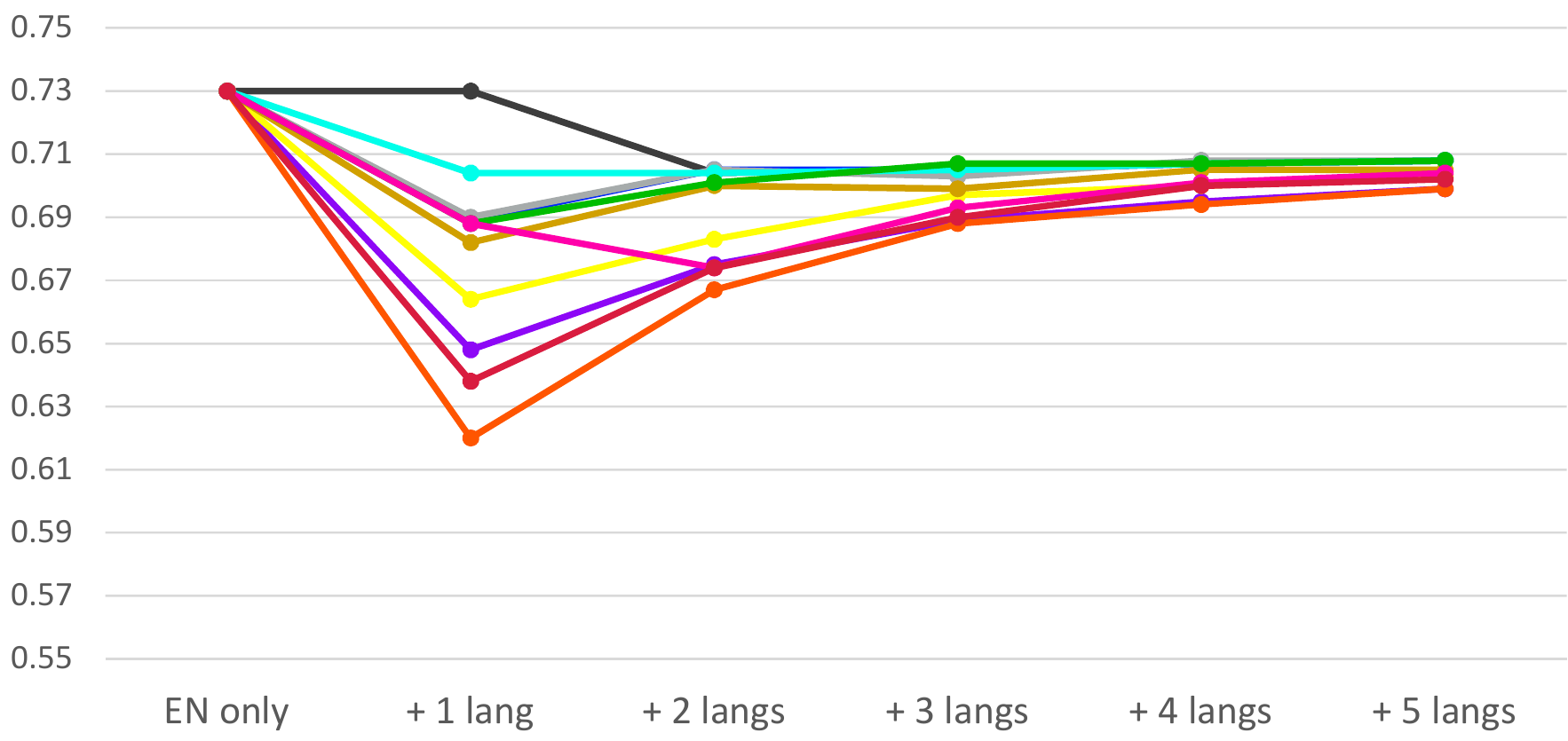}}
\subfigure[EN - Subtask~2]{\includegraphics[width=0.49\linewidth]{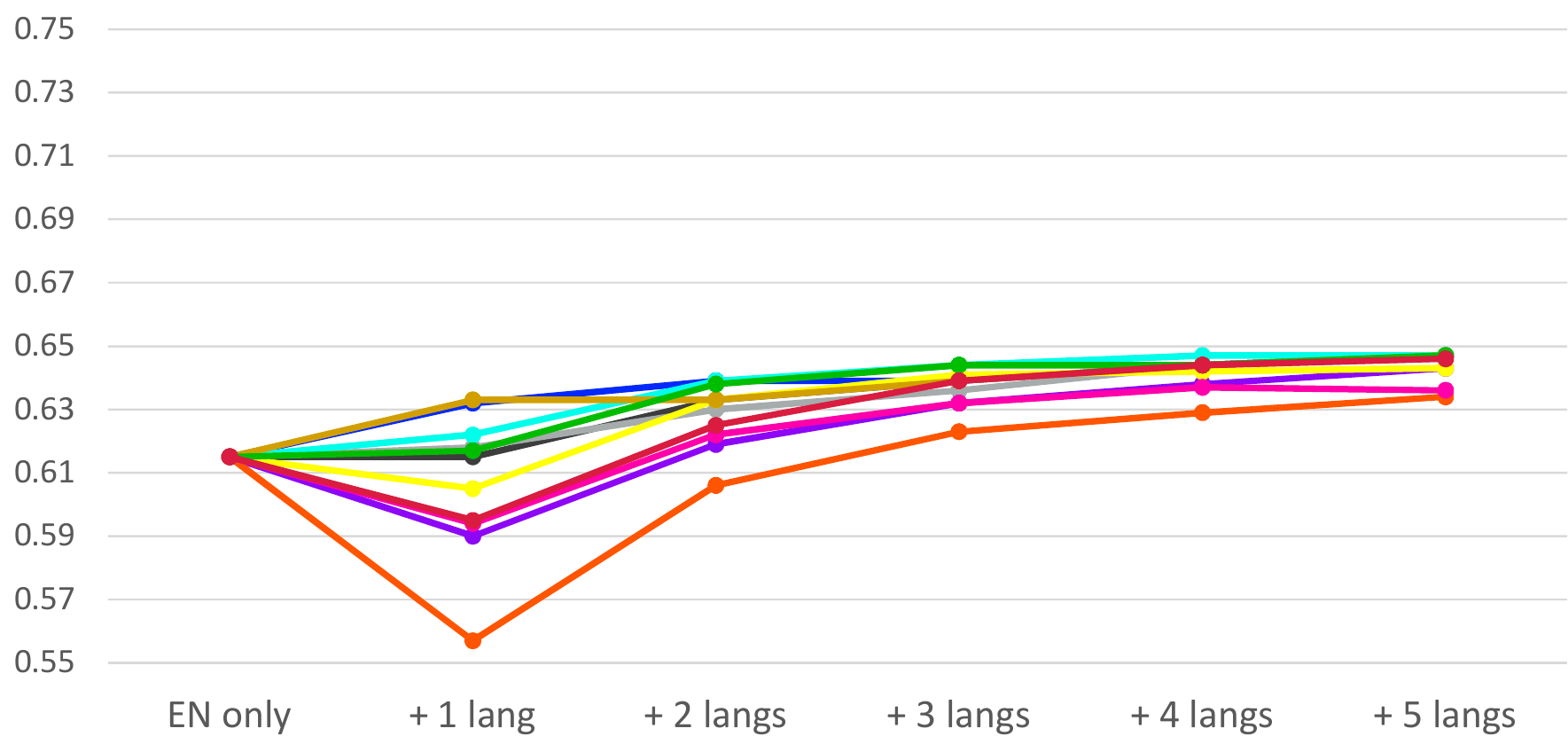}}\\
\subfigure[HR - Subtask~1]{\includegraphics[width=0.49\linewidth]{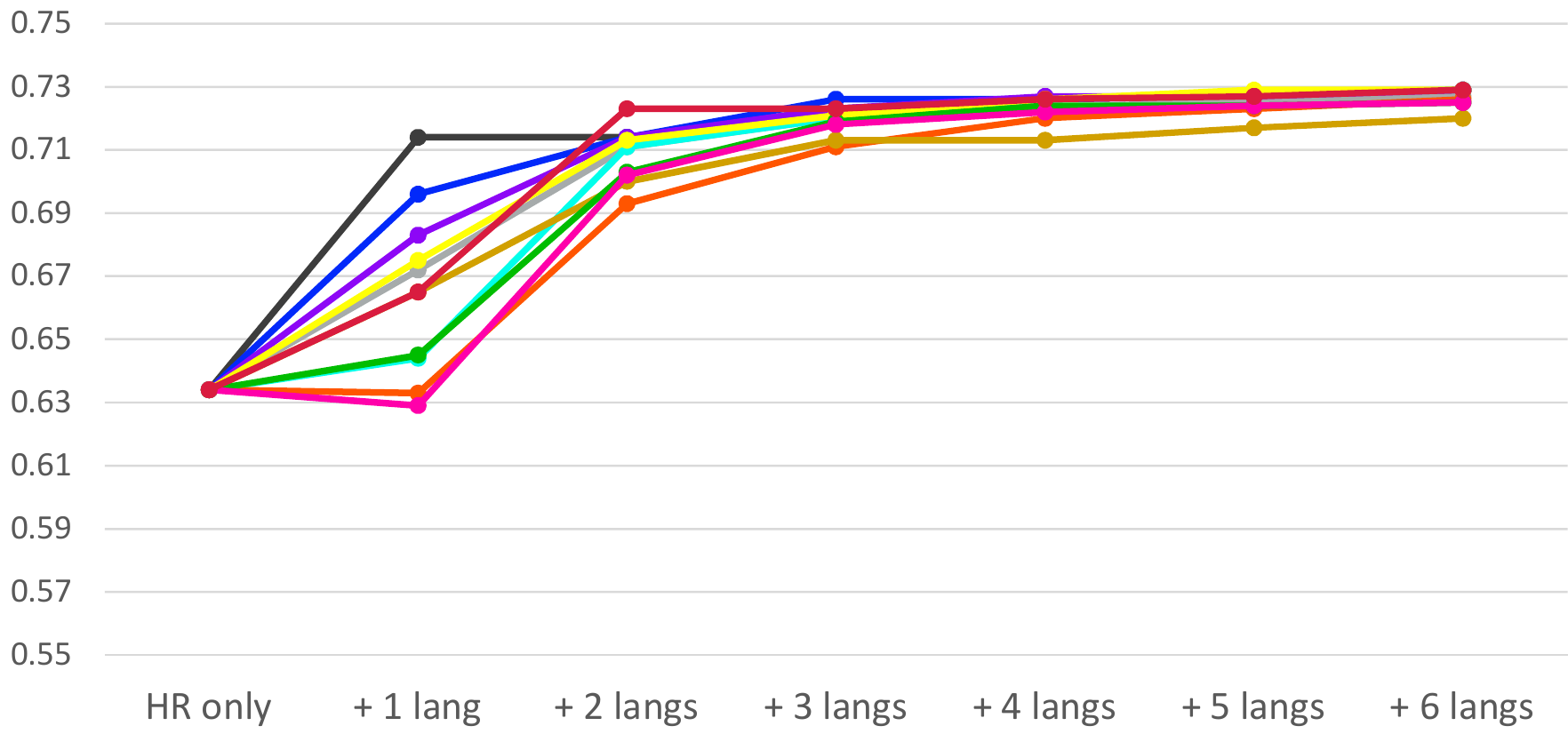}}
\subfigure[HR - Subtask~2]{\includegraphics[width=0.49\linewidth]{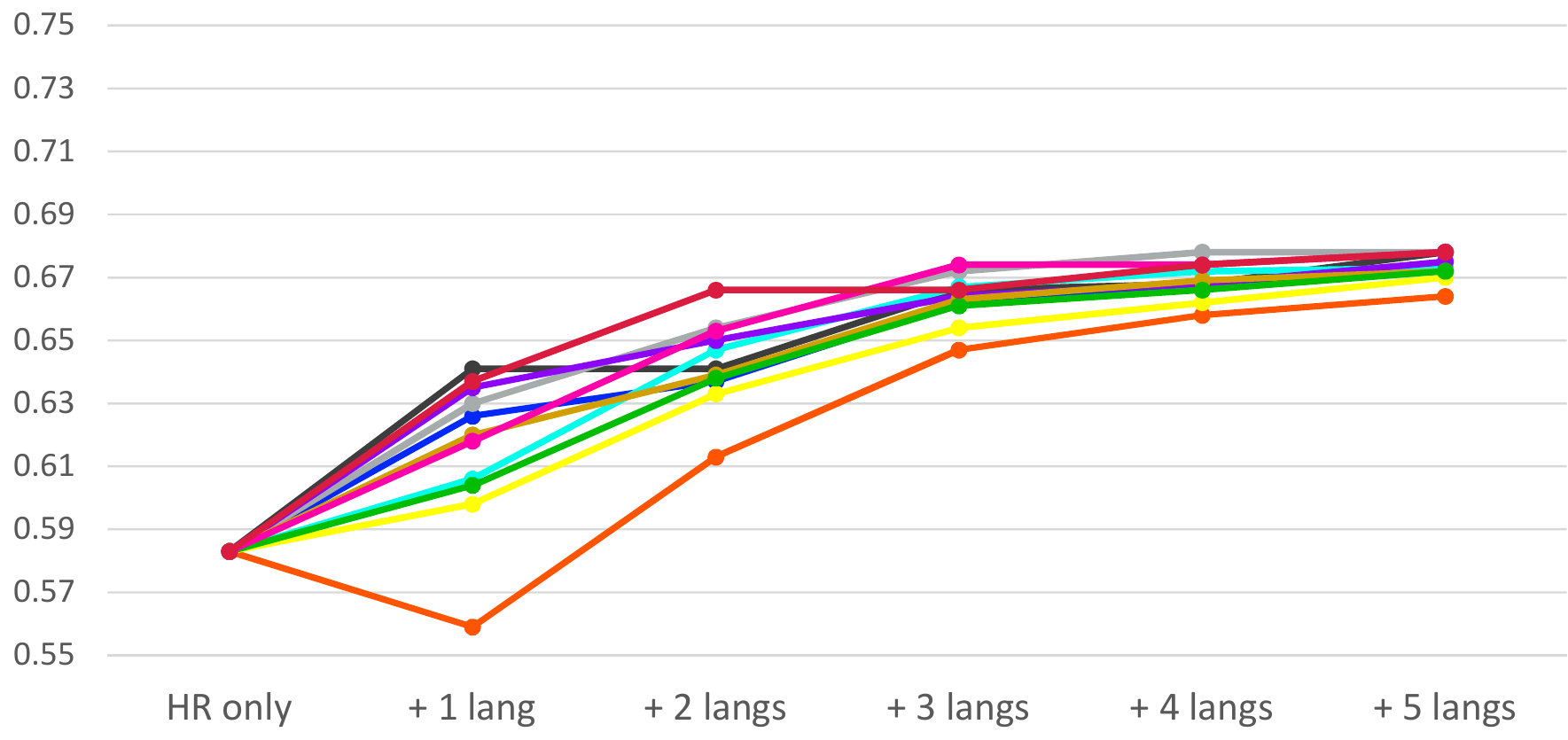}}\\
\caption{The effects of adding more languages to the similarity computation in \method. The numbers reflect the correlation with the human-generated similarity scores.}
\label{fig:lang_variation}
\end{figure}

\vspace{0.2cm}
\noindent\textbf{Do results improve as more languages are added? }
In order to evaluate the benefits of multilingualism, we performed an experiment in which the performance using only the source language (\ie the language of the original sentence) is compared to the performance when more languages are incrementally added. Figure~\ref{fig:lang_variation} shows the results for this experiment for the datasets in English and Croatian.
The first set of points on the plot mark the case in which only the original language was used.
The second set, shows the scores when each of 11 possible languages were added.
From the third set of points onward, we kept the language(s) that brought the biggest gain and added one more.
We repeated this process until the addition of a new language ceased to bring improvements. 
The combination of multiple languages was beneficial for Croatian, in both subtasks, and for English in Subtask~2. 
In Croatian, the addition of one language improved results in 9 out of 11 possible languages. The exceptions were Greek and Serbian, in which cases, the scores remained the same. 
By adding English, the score increased by eight percentage points. By adding further languages, the improvement was smaller but steady until it reached a plateau with six additional languages.

\begin{figure}[htb]
\centering
\includegraphics[width=0.9\linewidth]{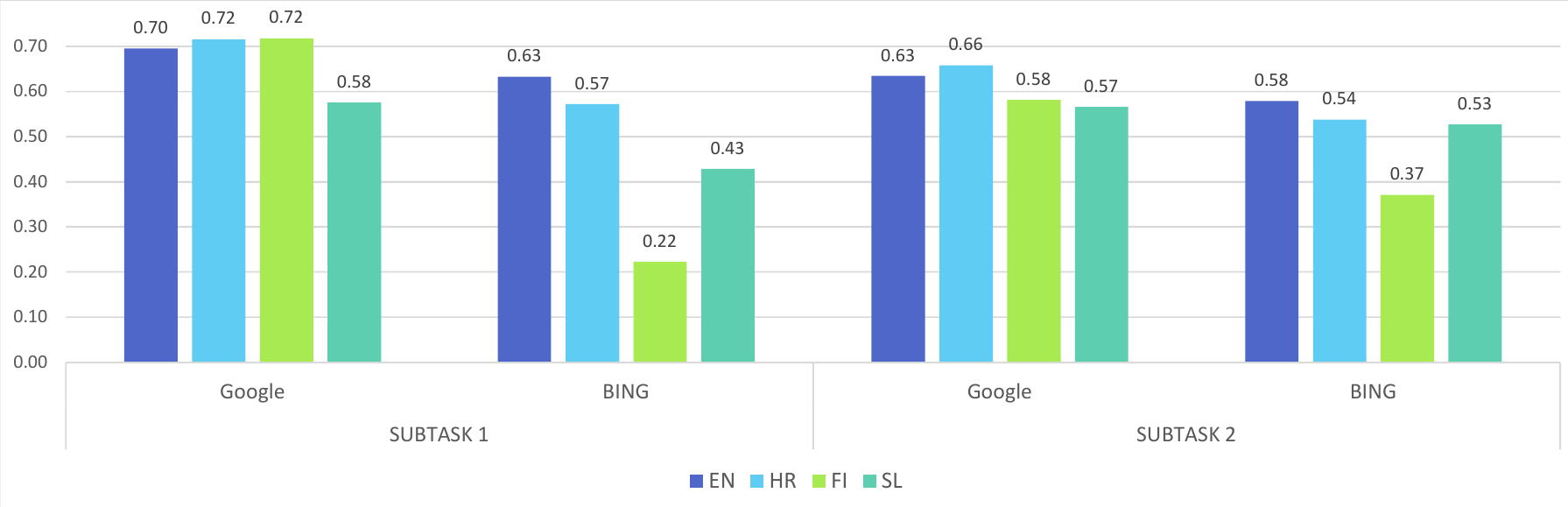}
\caption{Correlation scores with human judges for different translation engines in \method}
\label{fig:translation}
\end{figure}

\vspace{0.2cm}
\noindent\textbf{Does the translation mechanism impact the results?}
In order to evaluate the impact of different translation engines on \method, we compared the performance of Google Translator and Bing Microsoft Translator. The four original datasets were translated into the 11 languages using both engines. 
Then, the translated datasets were used to perform the contextual similarity tasks with the same algorithm configuration (all languages considered, $\alpha = 0.7$ and $\beta = 0.3$). 
The results are shown in Figure~\ref{fig:translation}. Google Translator outperforms Bing in both subtasks for all languages. 
The superior performance of Google Translator is in line with the findings from other recent works -- 
\newcite{marzouk2019} evaluated translations from of German to English and found that Google presented the best results and
\newcite{way2020} evaluated the translation of technical texts 
and found that, in most of cases, the translations provided by Google were better. 
Intuitively, better translations yield better contextual similarity and that was confirmed here.

\section{Conclusion}

In this paper, we described our system submitted to SemEval-2020 Task 3. 
We designed an approach that relies on translation and multilingual language models in order to compute the contextual similarity between pairs of words. The key idea is that having similarity information from different languages may help decide on how similar the words are. Our system achieved competitive results in both subtasks, being ranked among the top-3 in most runs.

In these preliminary experiments, we could not establish in which cases more languages are helpful and we leave it as future work.
Additionally, we are interested in understanding which factors contribute to improvement in the results -- whether it is the amount of data used for training the language models or individual features of the language.

\vspace{0.4cm}
\noindent\textbf{Acknowledgement.}
\small{This work was partially supported by CNPq/Brazil and by CAPES Finance Code 001.}

\bibliography{semeval2018}
\bibliographystyle{coling}

\end{document}